\documentclass[letterpaper, 10 pt, conference]{ieeeconf}  

\IEEEoverridecommandlockouts                              

\usepackage{graphicx} 
\usepackage[export]{adjustbox}
\usepackage{textcomp}
\usepackage{hyperref}

\newcommand{\customtilde}{\raisebox{0.15ex}{\texttildelow}}
\title{\LARGE \bf
The Interaction Flow Editor: A New Human-Robot Interaction Rapid Prototyping Interface
}

\author{Matthew Huggins$^{1}$, Anastasia K. Ostrowski$^{1}$, Andrew Rapo$^{1,2}$, \\Eric Woudenberg$^{2}$, Cynthia Breazeal$^{1,2}$, and Hae Won Park$^{1}$
\thanks{This work was supported by the UX Innovation Lab, Samsung Research.}
\thanks{$^{1}$Matthew Huggins, Anastasia K. Ostrowski, Andrew Rapo, Cynthia Breazeal, and Hae Won Park are with MIT Media Lab, Cambridge, MA, USA. {\tt\small \{hugginsm,akostrow,cynthiab,haewon\}@mit.edu, andrew.rapo@gmail.com}}%
\thanks{$^{2}$Andrew Rapo, Eric Woudenberg, and Cynthia Breazeal were with Jibo, Inc., Boston, MA, USA.}%
}

\begin{document}

\maketitle
\thispagestyle{empty}
\pagestyle{empty}

\begin{abstract}
Human-robot interaction can be regarded as a flow between users and robots. Designing good interaction flows takes a lot of effort and needs to be field tested. Unfortunately, the interaction flow design process is often very disjointed, with users experiencing prototypes, designers forming those prototypes, and developers implementing them as independent processes. In this paper, we present the Interaction Flow Editor (IFE), a new human-robot interaction prototyping tool that enables everyday users to create and modify their own interactions, while still providing a full suite of features that is powerful enough for developers and designers to create complex interactions. We also discuss the Flow Engine, a flexible and adaptable framework for executing robot interaction flows authors through the IFE. Finally, we present our case study results that demonstrates how older adults, aged 70 and above, can design and iterate interactions in real-time on a robot using the IFE.
\end{abstract}

\section{INTRODUCTION} \label{introduction-section}

As robots become more common in everyday settings and interact more frequently with the general population, it is essential that those interactions can be designed so that robots can collaborate effectively with humans, be integrated into human life without being obtrusive, and can best serve their functions in order to help people in their everyday lives. However, the interaction design process is often very disjointed, with users experiencing prototype interactions, designers forming those prototypes, and developers implementing them as mostly independent processes. A designer might interview a user, write up a specification of the interaction, and then give it to a developer to implement, before bringing the working prototype back to the user for feedback. Not only can this cycle be very slow, but it distances the user's actual experience of the prototype from the prototype's implementation, and makes real-time adjustments to the experience impossible.

Previous prototyping tools help remedy these issues by providing a graphical interface that allows a designer or other expert to quickly prototype robot interactions~\cite{interaction-composer}\cite{robostudio}\cite{choregraphe}. These tools provide various building blocks that the designer can combine to create a new interaction, and allow designers to quickly modify their prototypes in response to user feedback. However, many of these tools still require a level of expert knowledge that prevents an everyday person from modifying the interaction to their liking. Additionally, these tools often cannot support the full breadth of robot functionality that might be needed to create a complete interaction prototype, such as logical branches in the interaction, or rich human-robot dialogue.

In this paper, we present the Interaction Flow Editor (IFE), a new robot interaction rapid-prototyping tool that addresses these issues, by providing a spectrum of functionalities and user interfaces for novice to expert users to create, modify, and test robot interactions.  The IFE allows a single user to program, design, and experience a robot interaction all at once. While the IFE can easily be used by a novice, the suite of features it provides is powerful enough that it is also a great platform for developers and designers to build more complex, parallel programs. Such flexibility positions the IFE as a powerful rapid-prototyping interface that enables collaboration between researchers and users of the technology. 

The IFE provides a set of ``Flow Activity" blocks that can be combined to create interaction ``Flows". IFE is also robot-platform agnostic; with its flexible activity block design, it can easily adapt to work with any robot that passes its commands and messages via a web socket protocol, such as the bridge for Robot Operating System (ROS)~\cite{ros} as we demonstrate later in the paper.

The contributions we present in this paper are as follows:
\begin{itemize}
    \item The Interaction Flow Editor, a novel human-robot interaction prototyping tool that provides an unprecedented combination of ease-of-use and power to create rich social interactions.
    \item The Flow Engine, a flexible and adaptable system for interpreting the Flow and executing it on a physical robot.
    \item A case study that demonstrates the use of the Interaction Flow Editor in a participatory design study in which users were engaged in a rapid-prototyping participatory design session with design researchers to develop robot-interactions for older adults. 
\end{itemize}

In the next section, we highlight the key design goals of the IFE. Section~\ref{related-section} provides an overview of robot interaction design tools and their usage in participatory design approaches, as well as a discussion of their key strengths and shortcomings. Next, in Sections~\ref{editor-section} and~\ref{engine-section}, we describe the design of the IFE itself, the features it provides, and how the Flow can be executed on a physical robot by the Flow Engine. Finally, we discuss the use of the IFE in a study where older adults aged 71-86 developed their own robot experiences using the tool in Sections \ref{older-adults-section} and \ref{evaluation-section}.

\section{RELATED WORK} \label{related-section}


Many tools have been made to help experts, such as developers or designers, create and evaluate human-robot interactions. Perhaps the most popular tool for robot developers is the Robot Operating System (ROS)~\cite{ros}, which is a framework that aims to simplify the creation of complex interactions across a variety of robot platforms. However, the use of these development environments requires significant expert knowledge, including general programming skills. In order to solve this problem, many tools use graphical user interfaces to reduce the amount of coding required. These graphical tools exist both in research settings as well as in commercial applications. For example, Aldebaran Robotics' Choregraphe~\cite{choregraphe} provides a graphical environment for programming high level behaviors on its humanoid robots. However, the focus of the Choregraphe is much more on programming specific robot behaviors and  motor controls, rather than authoring a full interaction scenario with a user. The Interaction Composer presented by Glas et al.~\cite{interaction-composer} combines modular software with an easy-to-use graphical interface to allow programmers and interaction designers to develop robot applications together. While the Interaction Composer allows designers and developers to work together to create new interactions, RoboStudio~\cite{robostudio} allows users without expert knowledge to customize the behaviors of a robot to meet their needs.  Finally, Interaction Blocks~\cite{interaction-blocks} makes creating human-robot interactions easy by providing several interaction design patterns and rules that can be combined to create full interaction prototypes.


While these tools provide good interfaces for easily creating or modifying robot interactions, they each have one or more of the following three limitations (which mirror our three main design goals). First, no tool allows users without expert knowledge to build complete, fully functioning interactions. In previous systems, either an expert designer was required to build a complete interaction, or a user could only modify an existing interaction. Second, these systems are created for relatively narrow use cases. They are not designed to work with a wide range of robots, and to allow for external API integrations. Finally, while existing tools provide building blocks for rich interactions, they do not support multi-turn conversations powered by natural language processing with branching in the interaction.


\section{DESIGN GOALS} \label{goals-section}

Based on the review of prior work, we have identified the three key design goals of our Interaction Flow Editor (IFE), that can assist the participatory and co-design processes and provide users with the extended ability to program their personal robot companions:
\begin{itemize}
\item{Ease of Use:} Any person, with and without programming or robot design experience, should be able to create and modify their own Interaction Flows. This helps bridge the gaps between user, designer, and developer, by enabling anyone to assume all three roles. 
Ease-of-use is not only  important for novice users, but also for expert developers. For instance, programming branching interactions - like dialogue interactions - can be difficult, even for experienced coders. The simplest dialogue interactions have many branches and can require complex state to be managed. Typically, programmers use flow-chart-like diagrams to design branching interactions. Manually translating these diagrams into code is mentally taxing and inevitably error prone. It is difficult to ensure that the code faithfully represents the diagram. 

\item{Adaptable for Changing Platforms:} It is easy to expand the IFE with new Flow Activity blocks, allowing it to be used for a wide range of applications. The tool is robot-agnostic, meaning that it is trivial to adapt existing blocks to be used with a new robot, or when new hardware, software, and algorithms are added to the existing robot. 

\item{Rich Interactions:} The IFE is able to support rich and complex interactions, including different branches controlled by Flow logic, dialogue with both natural language understanding and response generation capabilities, and quick integration with external APIs or technologies.

\end{itemize}

\section{INTERACTION FLOW EDITOR DESIGN} \label{editor-section}

\begin{figure}[t]
      \centering
      \includegraphics[width=\columnwidth]{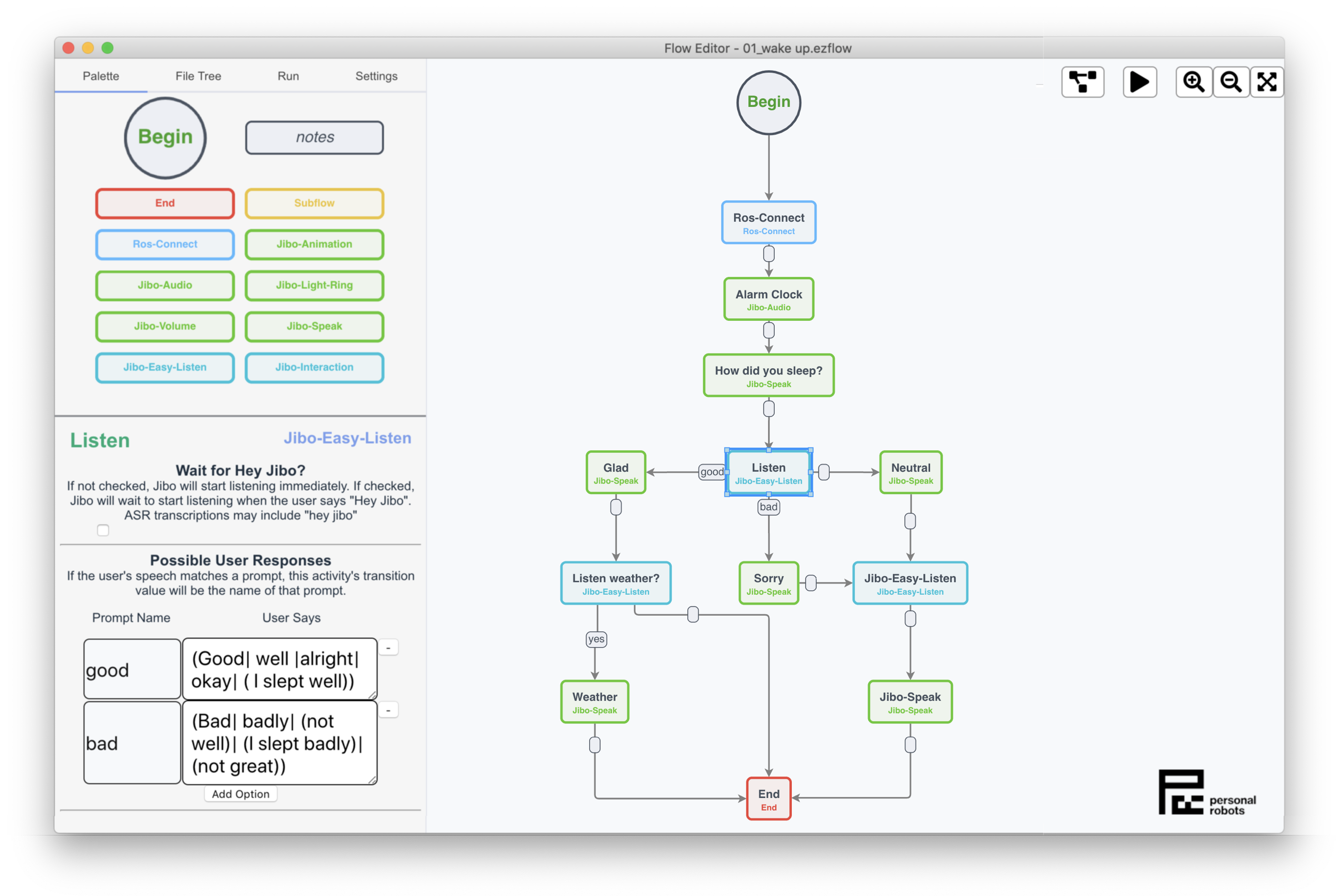}
      \caption{The main view of the Interaction Flow Editor.}
      \label{mainview-figure}
   \end{figure}

The IFE allows both a user with minimal previous experience to easily create and experience human-robot interactions and an expert user to develop complex interactions incredibly quickly. Each interaction is represented by a ``Flow", which is a graphical representation made up of several Flow Activity blocks connected by transition arrows. Once the user creates a Flow in the application, it can be executed on a physical robot using ROS.  
In this paper, we use Jibo robot to demonstrate Interaction Flow examples. Jibo is an 11-inch tall and 6-inch wide table-top social robot with a touchscreen face and three degree-of-freedom expressive body that provides interpersonal movements during an interaction\footnote{Jibo Robot, https://www.jibo.com/}. Jibo has microphone array on the front and back of its head, two cameras of which one has a wide angle  view , capacitive touch sensors along the back of its head, a stereo speaker, an accelerometer, and a proxemity sensor. It uses its custom engine for text-to-speech (TTS) and Google cloud speech-to-text service for automatic speech recognition (ASR). We developed a ROSbridge client for Jibo using the standard ROS JavaScript library\footnote{ROS JavaScript Library, \url{https://github.com/RobotWebTools/roslibjs}}.

\subsection{GRAPHICAL INTERFACE}

The IFE’s main view can be seen in Fig.~\ref{mainview-figure}. On the top left, there is a navigation menu, currently displaying the palette of Flow Activity blocks beneath it. On the bottom left, the options for the selected Flow Activity block (“Say Hi”) are shown. The majority of the screen is dedicated to the Flow Canvas on the right. The navigation menu has four tabs: Palette, File Tree, Run, and Settings. The Palette (at the top) displays all of the available Flow Activity blocks, which can be added to the Flow by dragging them onto the canvas. If an activity block on the canvas is selected (by clicking on it), the Palette also shows that activity’s options on the bottom. The File Tree tab shows the current project’s files and folders. 
The Flow Editor supports two types of Flow files, ``.flow” and ``.ezflow". When working with a ``.ezflow" file, the IFE provides a simplified experience that supports everything needed to prototype  high-level robot interactions, including Acitivities that support conversational turn-takes between the robot and the user, i.e., text-to-speech, play animation, listen, and parse user speech. The “.flow” file format supports the full extent of  low-level and advanced options, including custom synchronous and asynchronous JavaScript codes, external APIs, and more advanced Flow controls such as sub-Flows, parallel Flows, and exception handling. The advanced features are discussed in the next sections. Here we introduce the Flow Activities.

\subsection{FLOW ACTIVITY BLOCKS}

\begin{figure}[t]
      \centering
      \includegraphics[width=.7\columnwidth, frame]{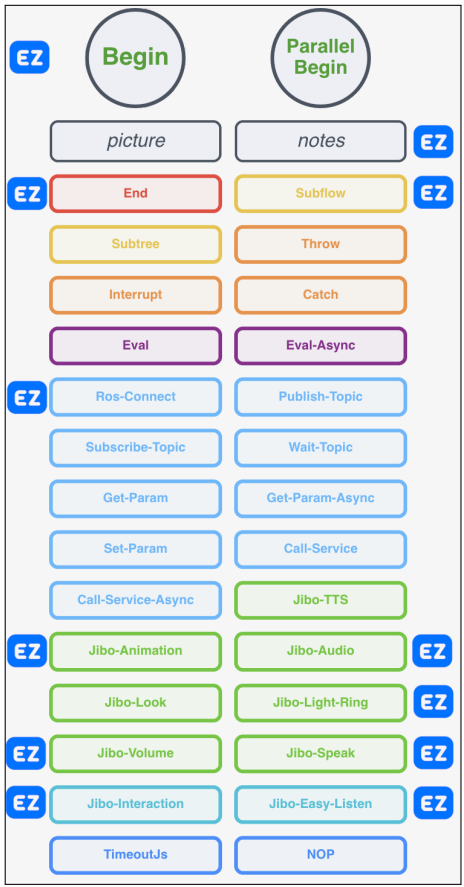}
      \caption{The available Flow Activity blocks. Blocks available in the simplified ``.ezflow" mode are market with the ``ez" icon.}
      \label{blocks-figure}
  \end{figure}

The IFE provides 32 different Flow Activity blocks that can be combined to create interactions (Fig.~\ref{blocks-figure}). Twelve of these make up the ``Easy Flow" set that is best suited for non-expert users to build event-based high-level interactions. By connecting Flow Activity blocks with transitions, the user can build a full interaction Flow, with dialogue between the robot and the user being the main part of the interaction. Each Flow Activity block takes optional inputs, for example, the Speak block allows the user to provide text for what the robot Jibo will say. 

Flow Activities fall into several categories, including general and advanced functionality Flow  blocks such as begin/end, exception handling, timeout, etc., a full suite of ROS Activity blocks that are needed to interface with ROSbridge, and lastly, Robot Activity blocks that can be adapted for each robot-platform dependent activities (here we show Jibo blocks as an example). Robot Activity blocks are compiled at run-time and take as input ROS message topic name and message definitions, thereby can be easily adapted to different robot platforms' existing ROS node and message network. Detailed descriptions of the available Flow Activity blocks can be found below in Section~\ref{ACTIVITY-BLOCKS}.

\subsection{TRANSITIONS BETWEEN FLOW ACTIVITIES}
In a Flow, individual Flow Activity blocks are connected by transition arrows. To add a transition, the user can click on the first block and drag the arrow to the next. During execution, once a Flow Activity has finished (e.g. Jibo has finished listening to the user, or Jibo has finished an animation), execution proceeds along a transition to the next Flow Activity to start. Some blocks, such as Jibo-Easy-Listen and Jibo-Interaction, can have multiple outbound transitions in order to allow branches in the interaction. The transition chosen during execution is found by comparing the Activity's ``result" property (i.e. its output value) to the names of the Activity's various outbound transitions. If a match is found, that transition is taken. If there is no match, the unnamed transition will be used -- here acting as the “else” path when there are multiple outbound transitions. More details on how a Flow is executed at run time can be found below in Section~\ref{engine-section}.

\subsection{RUNNING INTERACTIONS WITH A ROBOT}

Once a Flow has been created, the user can execute the Flow by pressing the play button in the upper-right hand corner of the application window. The Flow Engine loads the Flow from a file and begins execution. If the Flow contains a ROS Connect Activity, the Flow Engine will connect to ROSbridge using the IP address provided in the block's options. From then on, the Flow Activities can be executed on the robot via the robot's ROS network. The Flow Engine is discussed in detail in Section~\ref{engine-section}.

\section{FLOW ACTIVITIES} \label{ACTIVITY-BLOCKS}

\subsection{FLOW FUNCTIONALITY}

Every Flow must start with a Begin block, where the Flow Engine will start execution. A Parallel Begin block is also available, which allows for multiple Flows to run concurrently, e.g., one Flow is responsible for delivering dialogue interaction while a parallel flow controls the robot to track and attend to the user's face. Similarly, a Flow must always terminate with an End block. 

\subsection{ROS ACTIVITIES}
The ROS Activities cover the full range of ROSbridge functionalities, and can be used to control any ROS-enabled robot. They can also be easily expanded to create higher-level Flow Activities for a specific robot, as they are for our Robot Activities below. The most important ROS Activity is ROS-Connect, which allows the Flow to connect with ROSbridge, enabling future communication with the robot.

In order to send a message to a robot or other ROS service, the Publish-Topic activity can be used. The Flow can wait for a message using Wait-Topic, or set a callback to be executed whenever a message is received using Subscribe-Topic.

ROS parameters, i.e., topic name and message definition,  can be set with the Set-Param activity, and retrieved either synchronously or asynchronously with Get-Param and Get-Param-Async, respectively. Finally, a ROS service can be called, again either synchronously or asynchronously, using Call-Service or Call-Service-Async. 

\subsection{ROBOT ACTIVITIES}
\label{sec:robot-activities}
The Robot Activities support higher-level commands that are extensions of the ROS Activities. In the Flow Engine, they use the Publish-Topic and Wait-Topic Activities to send commands and receive feedback from the robot. While we use these Activities with Jibo, they could easily be adapted for any other robot by implementing a simple receiver skill on the robot that translates the commands into actions.

The most valuable Activities provided by the IFE are those that enable dialogue between Jibo and the user. There are two Activities that allow Jibo to speak. The Jibo-Speak Activity provides a simple interface, where the user can type in one or more options for what Jibo will say (if multiple options are provided one will be chosen at random). The Jibo-TTS Activity allows the user to provide a JavaScript function to return a string that Jibo will say. 

\begin{figure}[t]
      \centering
      \includegraphics[width=.6\columnwidth, frame]{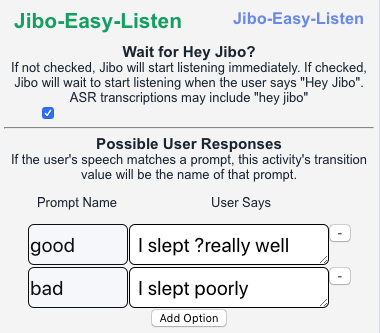}
      \caption{Example options for the Jibo-Easy-Listen block. After Jibo listens to the user speak, a rule-based grammar parser will attempt to match the ``User Says" values against the transcription of the user's speech. If a transition leaving the block has the same label as the matching prompt name, the Flow Engine will take that path.}
      \label{listenoptions-figure}
  \end{figure}

For Jibo to listen to the user, and to change the course of the interaction based on what the user says, Jibo-Easy-Listen can be used. An example of the options for the Jibo-Easy-Listen block can be seen in Fig.~\ref{listenoptions-figure}. First, the user can choose whether Jibo will start listening immediately when the Activity is executed, or if he will wait for the wake word (``Hey Jibo") before listening. Once the user is done speaking, Jibo's ASR sends a transcription of the user's speech back to the Flow Engine. The Flow Engine then tries to match the transcription against one of any number of prompts, provided they are under ``User Says" in the options view. These prompts can use rule syntaxes that is compiled by a rule-based grammar parser to extend the phrase matching. In the example in Fig.~\ref{listenoptions-figure}, the ``good" prompt can match either ``I slept well" or ``I slept really well". For expert users, a rule file can be created to define a set of rules and build a more complex parsing system. The rule-based grammar parser provides a rich suite of options for creating complex rules, including optional phrases, either-or options, wildcards, and pre-defined sub-rules. These features allow users to create rules with a similar amount of complexity to a typical context-free grammar parser used in industry applications \cite{slu}. If a match is made between the ASR result and one of the prompts, the Flow Engine will use the name of the prompt as the transition value for moving to the next Activity in the Flow. In our example, if the user says ``I slept poorly", the Flow Engine will look to follow a transition from the Jibo-Easy-Listen activity with the label ``bad". In this way, multi-turn dialogues can be created. 

There is also a Jibo-Interaction Activity, which combines Jibo-Speak and Jibo-Easy-Listen with additional ``No Parse" or ``No Speech" backup responses for Jibo. With just a single Jibo-Interaction block, Jibo can prompt the user, listen for a response, ask the user again if the response was not understood, and decide what do do next all in one activity. The depth of this functionality is consistent with deployed voice agents and chatbots, and provides a powerful prototyping tool.

Along with Flow Activities for dialogue, there are several other Robot Activities that together allow for an expressive interaction of the robot, such as the animation, audio (play sound), volume, look-at (orient attention to a given point in the 3D space), and light-ring (change the color of an LED ring) blocks. 


\subsection{ADVANCED FLOW CONTROLS}
While the ROS and Robot Activities provide an interface for controlling robots, there are additional Flow Activities available for more advanced functionality, including  exception handling, and encapsulating existing Flows. The Eval and Eval-Async Activities can be used by developers to execute synchronous and asynchronous JavaScript code. This allows for external APIs to be used within a Flow, and for nearly any otherwise unsupported functionalities. The Throw, Interrupt, and Catch Activities provide full exception handling logic within Flows (these are discussed further in Section~\ref{EXCEPTION-HANDLING}). Flows saved in other files can be loaded as a single SubFlow Activity to be used as a component in a new Flow. Finally, there is a TimeoutJS Activity that pauses for a given number of milliseconds, and a NOP (``no operation") Activity that can be used for organizational purposes (e.g., joining branches).

\section{THE FLOW ENGINE} \label{engine-section}

While the Flow Editor provides an interface for creating, modifying, and testing Flows, the Flow Engine is responsible for executing them. 

\subsection{FLOW PROGRAM EXECUTION}

A Flow contains individual Flow Activities linked together by transitions. The core unit of a Flow is a Flow Activity. Each Flow Activity is a step in the overall Flow, and is executed independently of the other Activities.  
When executing a Flow, the Flow Engine first creates a Context for the Flow and then starts execution at the Flow's \textit{Begin} Activity. A Context can be thought of as the stack frame in which a Flow is executing. It holds the current Activity pointer (the ``PC") and the local variables (the ``notepad") for an executing procedure. Whenever a Flow invokes another Flow (a sub-Flow), the existing Context is pushed on the stack and a new Context is created in which the invoked Flow will execute. When the sub-Flow completes, the results are made available to the calling Flow. In this way, the Flow programming model resembles a typical stack-oriented programming language.

Upon completion of any Activity, the Flow Executor will take the just-executed Activity's result (converted to a string) as the name of the transition to follow. If none of the named transitions match the result, the unnamed transition is followed. If no unnamed transition exists and no matching transition is found, a Flow Engine exception is thrown. An example of transition execution is provided in Section~\ref{sec:robot-activities}.

The execution of a Flow is driven by calling each Activity's \textit{start} and \textit{update} methods. The start method is called once to perform any initialization the Activity requires. The Flow Executor then repeatedly calls the update() method until its return value indicates that the activity is complete, at which point the Flow Executor transitions to the next Activity. The frequency - or frame rate - of this repeated call to update depends on the nature of the application and is typically between 10FPS and 30FPS.

Flow Activities are implemented in JavaScript (JS) and can be extended with custom JavaScript code. Any JavaScript code run on behalf of the activity (that is, the JS code in the start and update methods) have access to 3 JS objects: the \textit{environment}, the \textit{blackboard}, and the \textit{notepad}. The environment is a collection of miscellaneous read-only information related to the Flow and FlowExecutor. It is intended to allow a Flow to adapt to variable aspects of its run-time environment. The blackboard is a single object that is created when the Flow starts and is intended for holding global variables, or information that must be passed between disparate parts of Flow. The notepad is an object within the Context object that is created whenever a Flow begins execution. If a Flow calls itself recursively, a new notepad is created upon every invocation. It represents the private local variables for that invocation of the procedure.

\subsection{EXCEPTION HANDLING} \label {EXCEPTION-HANDLING}

If an exception is thrown, either by a Throw Activity, or by JS code running on behalf of a Flow Activity, the Activity’s result is set to the name of the exception, preceded by a tilde (e.g., ``\customtilde Human.InteractionError.noInput"). The Flow Engine uses an iterative name-matching and activity-searching process to find the next activity to execute after an exception occurs. If there is no match, the exception name is shorted by successively dropping the suffix. In this way, a single Catch activity named \texttt{\customtilde} on the top page of a Flow Program can be used to catch all otherwise uncaught exceptions occurring within the program.

\subsection{EXTENDING THE FLOW ENGINE WITH CUSTOM ACTIVITIES}


The Flow Engine provides an easy way to define new, custom Activities. The FlowExecutorFactory has a register method that allows new classes to be dynamically added to the ActivityClassMap. Once added, Activities can be referenced by name in a .flow file and then instantiated at run time by looking up this name in the map. 


\subsection{FLOW ENGINE PERFORMANCE}

There is overhead associated with the mechanism used by the Flow Engine to dynamically look up and instantiate Activity classes as they are encountered at run time. For this reason, Activities should be used in low-frame-rate loops. Tighter, computation-intensive loops should be hand coded and/or encapsulated in a specialized Activity.

\section{A CASE STUDY: RAPID-PROTOTYPING WITH OLDER ADULTS} \label{older-adults-section}
In this case study, we sought to explore how the Interaction Flow Editor (IFE) aided in the process of co-designing social robots with older adult users. Our participants engaged in a rapid-prototyping session in which they designed how they want the robot to interact with them in various daily events.  

Inviting older adults in the robot design process is not new. Our and others' previous works implemented various participatory design approaches to engage older adults in creating robots through interviews, design workshops, storyboarding, and low-fidelity prototyping~\cite{rose2017designing,ostrowski2019older,lee2017steps}. However, none of these studies attempted to involve older adults in the programming of robot skills and interactions. If one cannot prototype and iterate the interaction scenarios while experiencing the changes on the robot they are developing  the interactions for in real time, users and designers have to rely on their imaginations of what they want or see changes in the robot stagnantly between meetings with researchers, which were are limitations of prior robot participatory design approaches. Electronic HCI toolkits have been developed  aiming to enable older adults to lead participatory design processes~\cite{ambe2019older, jelen2019older}. To our knowledge, this same attention has not been given to older adults leading the development of social robot interactions. 
We present the IFE as a key toolkit that can enable participants such as older adults to take a lead role in robot interaction development, and our case study  aimed at evaluating that.

\subsection{STUDY DESIGN}
Prior to the robot rapid-prototyping session, we sent Jibos to our participants' homes for a month. While living with the robot, the participants experienced its basic features, e.g., simple chit-chat, information retrieval such as weather and Wikipedia search, radio and dance functions.  This provided them enough time to understand how to interact with the robot, when it listens and parses users' speech, and what they do and do not like about the interactions. Afterwards, the participants were given a script of interaction scenarios that they edited and returned to the researchers. The interaction scenarios included various events that the robot could support during a typical day, ranging from schedule reminders to supporting connecting with others to emotional wellness to medication adherence, etc. Based on this input, our interaction design researchers made customized Flow templates for each participant with the IFE. Each interaction was edited in separately with a total of 14 interactions and, therefore, 14 Flow templates per participant. During the rapid-prototyping session, participants first created an example Flow to become familiar with the editor, then using the template provided by the researchers, created and iterated each of the 14 Flows while witnessing their Flow embodied on the real robot.

\subsection{PARTICIPANTS}
Eight older adults (average age = 77.3; range = 71 - 86; 6 women) participated in the study. Older adults were defined as older than 70 years of age for the context of this study. All but one participant was retired. Two participants had some programming experience in their careers using punch cards. Therefore, all participants were considered to be non-experts. Each participant completed an approved IRB consent form. No incentives were offered.

\subsection{DATA COLLECTION and ANALYSIS}
Data was collected through video, computer screen, and audio recordings. Audio recordings were transcribed and reviewed for accuracy. 

\subsubsection{QUANTITATIVE ANALYSIS}
The computer screen recordings were annotated for significant events, including every time a Flow Activity block was added, modified, or deleted by the participant, every time the researcher stepped in to fix an error in the participant's Flow, as well as the duration of the participant creating the example Flow and modifying any of the 14 interactions. The duration of modifications to a Flow was defined as the time from the beginning of the first modification through the finish of the last modification, excluding large periods of time spent discussing the interaction without using the tool. Annotation was done using ELAN~\cite{elan}, and was split evenly between two researchers, with one participant's session annotated by both researchers in order to assess agreement. After annotation was completed, the two researchers discussed any discrepancies until complete agreement was reached.

\subsubsection{QUALITATIVE ANALYSIS}
From the qualitative transcripts, researchers identified questions the participants asked during the session and participants' reflections as key parts of the session to analyze due to their benefit to understand participants' understanding and suggestions for the IFE and perceived ease-of-use of the system. The quotes and participant reflections were isolated and coded through a Grounded Theory approach~\cite{charmaz2014constructing} to identify themes present in the text. Emergent themes were discussed by the first, second, and final authors. The second author coded the transcripts with the themes based on this discussion and the coding was reviewed by the first and final authors. The first, second, and final authors discussed the thematic coding, resolving any discrepancies until agreement was reached. 

\section{EVALUATION} \label{evaluation-section}
\subsection{QUANTITATIVE USAGE ANALYSIS}
\label{quant-results}


   
\begin{figure}[t]
      \centering
      \includegraphics[width=8cm]{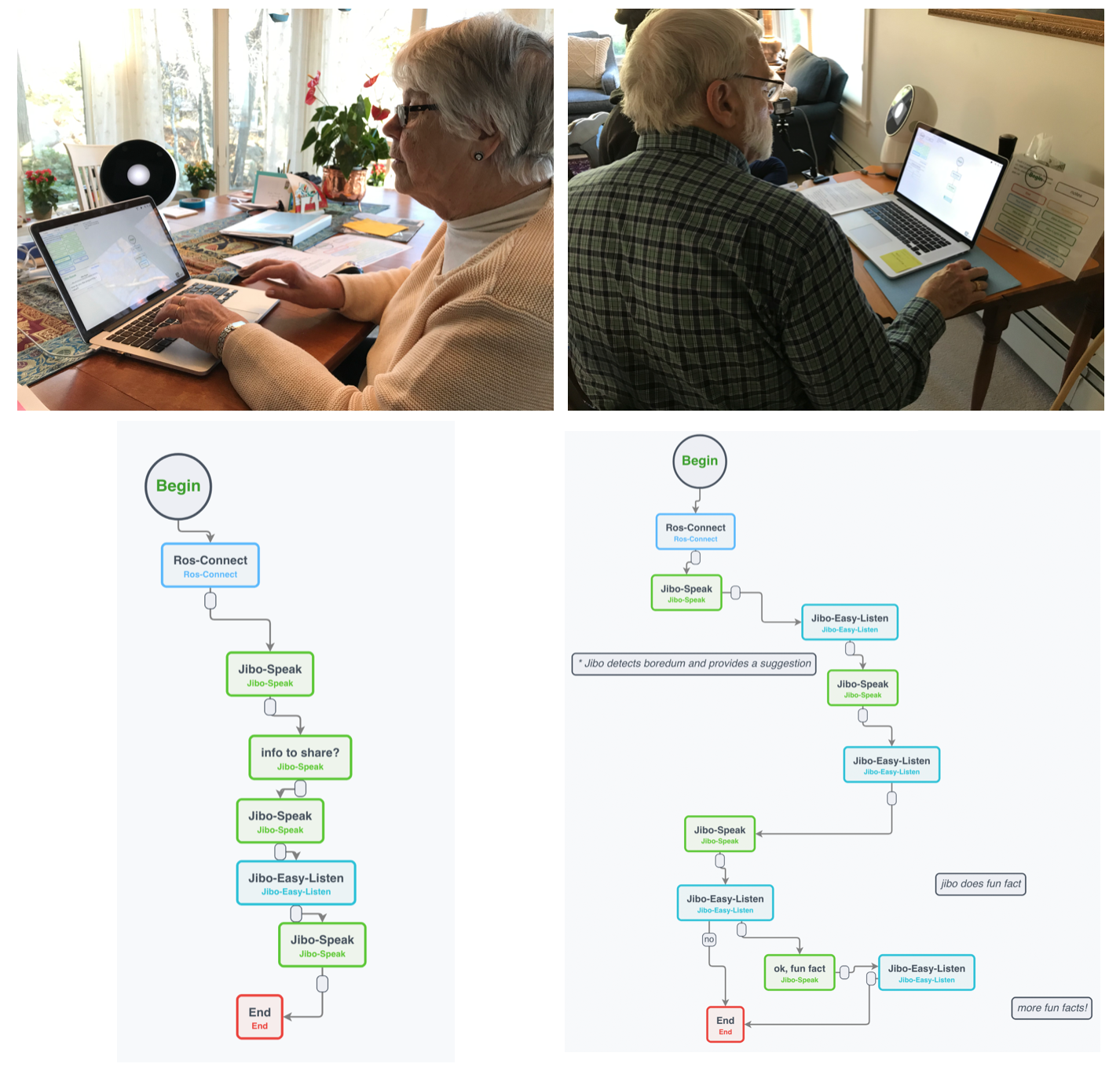}
      \caption{Older adult participants created and tested the Interaction Flows for various daily tasks with Jibo. Left: They created a check-in after going to  a doctor's appointment. Jibo plays subtle music in the background, listens for user's wake-up cue, and discusses how her sleep was. Right: They designed Jibo to check-in with the user in the afternoon and engage the user in fun facts.}
      \label{study-example-figure}
\end{figure}

\subsubsection{TUTORIAL}
At the beginning of their sessions, the participants were given a tutorial to create a Flow on their own in order to become familiar with using the IFE. An example Flow of the interaction created by a participant can be seen in Fig.~ \ref{study-example-figure}, although the final Flows created by the participants varied slightly. Four of the eight participants started with an empty Flow canvas, and four began with a template that had the Begin and ROS Connect blocks already added by the researcher. In all cases, the researcher set the ROSbridge IP Address option in the ROS Connect block for the participant. On average, it took 6 minutes and 48 seconds for the participant to complete the example Flow (standard deviation 4'13", range [2'23", 15'01"]). This results in just over a minute spent adding and configuring each block on average (mean 1'16" $\pm$ 48"). For two of the eight participants, the researcher had to step in and fix an issue with the participant's Flow (a single fix for one participant, two fixes for the other).

\subsubsection{SCENARIO EDITING}

After the example Flows, the participants went through 14 interaction scenarios, and were given the chance to make modifications as they saw fit. On average, each participant modified 7 of the 14 interactions (range [5,9]), and decided to delete 2 of interactions entirely (range [0,9]). In 25 of the 56 modified scenarios, the participant only made a single change (defined as adding or deleting a block, or modifying the options of a single block). These change took on average 37 seconds (standard deviation 24.6s). In the remaining 31 Flows, larger modifications were done, such as adding and modifying multiple new blocks or revising existing ones. Most frequent changes to existing blocks were revising robot speech and Listen result parsing options. Out of the 8 participants, 5 asked a question for clarification on how to use the system during the sessions (two participants asked a single time, two asked twice, and one asked four times). The researcher stepped in to fix issues with the participant's modified Flows in 7 out of the 56 modified scenarios. 




\subsection{QUALITATIVE ANALYSIS} \label{qual-results}
Participants' questions while using the editor circulated around the functions of the editor related to saving Flows and using the blocks. Specifically with regards to the blocks, participants asked about the blocks  and their activity meanings (i.e. \textit{``Easy listen is a radio program, right''} P05), connecting blocks (i.e. \textit{``You have to make a connection though, right?''} P08), and moving blocks (i.e. \textit{``So can I move [the block] up further?''} P08). 
Lastly, participants inquired about how the Flow connects to the robot performing the interaction. When first navigating playing the interaction on the robot, P08 asked, \textit{``Is that the Jibo interaction?''}, and P30 asked, \textit{``So then, where's respond? Interaction?''}.

While participants' questions serve as a lens to understand where confusion was occurring during the first time usage of the tool, participants vocalized the changes that can be made to the Flow Editor to improve interaction quality during their reflections about the tool. Seventy-five percent of participants mentioned an aspect of the tool in their reflections. P23, P20, and P13 mentioned the ease-of-use of the system. For example, P13 remarked, 
\begin{quote}
    ``I was pleasantly surprised that it was much easier than I expected...it's not something that comes natural to me. And I think the older you get and the less you use that kind of skill, it becomes harder. But this, I found I could play with this for a couple of hours without getting bored and, and have him do different things, and create a dialogue on Jibo that I have fun with.'' - P13
\end{quote}
P20 also commented that \textit{``It's easy. To me it's just a flowchart. I can deal with flowcharts.''}. There were some people that expressed some difficulty at initially getting started with the tool: \textit{``I had difficulty understanding how to, uh, get an interaction going.''} (P05). P25 and P08 commented on the tool's intuitiveness: \textit{``[The tool]...put it into a format that is concrete, so I kind of can understand what it's doing. I like to see how it's programmed.''} P08 added a note saying, \textit{``Couldn't have done it without you though.''}, demonstrating the importance to P08 of having some guidance with the tool, even though they were operating it. 
Participants also commented on the benefits of using the Flow Editor in conjunction with the robot. P13 highlighted, \textit{``It was nice to have some kind of tool where I could make it better or change some of the limitations that I found [in the robot]. So, this was a nice sort of exercise. I like it."} The interaction with the Flow editor provided participants with a deeper understanding of how the robot was functioning and being programmed (mentioned by 75\% of participants). Commenting on the benefit of understanding, P23 said, \textit{``It took some of the mystery out of it...it gave me a little bit of insight into the whole, into this whole area of programming...And it's clearly quite interesting...I've always thought...programming must be very boring, but this was quite interesting.''} Overall, participants were very proud of their Flow interactions they developed and contributing to the process of robot design. As P31 said, \textit{``This really makes me feel being a part of this whole study and I appreciate that, um, that our opinions matter and, um, that, uh, even though we're not programmers...we were still able to do some simple programming and I think that's fun.''} 


\section{DISCUSSIONS} \label{discussions-section}

The Interaction Flow Editor (IFE) is an easy-to-use yet powerful tool that allows complex human-robot interactions to be easily authored. By providing both a simplified ``.ezflow" mode and a fully-featured advanced ``.flow`` mode, the IFE can both create an authoring environment that is appropriate for a novice user and provide a fully-featured interaction development environment for designers and developers. Not only does the Flow Engine enable the intuitive block-based graph of a Flow, but it also allows for the system to be easily expanded to be used with new robots, to support new features, and for Flows to be deployed into production settings. This combination of an easy-to-use interface with a broad set of production-ready tools makes the IFE a powerful platform for both designing and developing human-robot interactions.

In our study, the large number of edits that participants made to the template Flows, which were designed from their initial input on the scenarios, show that experiencing the interaction on a physical robot and iterating the design in real-time is crucial in the design process. Not only were half of the scenario Flows edited by the participants, some others were completely deleted, as the participants realized that the initial design from their imagined scenario did not match their ideal experience with the robot. This demonstrates that the process of designers gathering feedback from users on written interactions may miss important feedback that can be discovered when the users experience the interactions with a physical robot. The IFE enables this process, providing a quick and easy way for users to experience the interaction and iterate on it. 


Overall, participants highlighted (1) the ease-of-use of the IFE, (2) its intuitiveness, (3) the value of being able to play an interaction on an embodied robot, and (4) personal growth. The first two points positively suggest that the IFE can be incorporated into participatory design activities with non-expert participants. The third point supports the importance of having participants be able to experience robot interactions live in rapid prototyping settings. The last point emphasizes the benefits of incorporating participatory design practices in human-robot interaction. In participatory design, researchers and participants engage in mutual learning where ``participants increase their knowledge and understandings: about technology for the users, about users and their practice for designers, and all participants learn more about technology design.''~\cite{simonsen2012routledge}  While there is a prevailing stereotype that older adults cannot master technology~\cite{durick2013dispelling}, and, therefore, cannot develop technology~\cite{ambe2019older}, participants in our case study proved it wrong. As both designers and users with the tool, they learned to use the IFE and, in the process, obtained knowledge on how to better articulate their ideal robot interactions. 

The case study and participants' subsequent feedback enabled developers to understand limitations of the IFE and places for improvements. Participants revealed areas of limitations including making the controls of the Flow clearer (i.e. clearer cursor, easier to connect blocks). One participant voiced that he could only use the tool in conjunction with researchers. This can be seen as negative or positive. It's important that this tool can be used in conjunction with researchers to promote participatory design, co-design practices, and mutual learning~\cite{lee2017steps}. However, the tool should also be intuitive that the participant could use it on their own without the help of the researcher, promoting a more equal power dynamic between users and researchers~\cite{lee2017steps}. Our mixed-method analysis approach reveals participants' usage patterns of the IFE, elaborates upon the importance of understanding participants' perception and opinions of the tool, expands upon the value of engaging researchers and participants in participatory design and co-design processes enabled by tools such as the IFE.

\section{CONCLUSION}
\label{conclusions-section}
In this paper, we explored how the Interaction Flow Editor enables both expert and novice users to create and iterate on social human-robot interactions by providing a set of Flow Activity blocks that cover a wide spectrum of functionalities. We discussed the Flow Engine, which is a powerful platform for executing Flows, and can easily be expanded with additional functionalities and for new robots. We conducted participatory design study with older adults, which gave us insight into how the IFE can be used to help participants and designers collaborate on human-robot interactions, and insight into how the tool can be even more easily used by novice users. Our mixed-method analyses show that the Interaction Flow Editor is a valuable tool that not only makes prototyping interactions easy for experts, but empowers non-expert robot users to be involved in the creation of those interactions. 




\bibliographystyle{IEEEtran}
\bibliography{citations}

\begin{thebibliography}{10}
\providecommand{\url}[1]{#1}
\csname url@samestyle\endcsname
\providecommand{\newblock}{\relax}
\providecommand{\bibinfo}[2]{#2}
\providecommand{\BIBentrySTDinterwordspacing}{\spaceskip=0pt\relax}
\providecommand{\BIBentryALTinterwordstretchfactor}{4}
\providecommand{\BIBentryALTinterwordspacing}{\spaceskip=\fontdimen2\font plus
\BIBentryALTinterwordstretchfactor\fontdimen3\font minus
  \fontdimen4\font\relax}
\providecommand{\BIBforeignlanguage}[2]{{%
\expandafter\ifx\csname l@#1\endcsname\relax
\typeout{** WARNING: IEEEtran.bst: No hyphenation pattern has been}%
\typeout{** loaded for the language `#1'. Using the pattern for}%
\typeout{** the default language instead.}%
\else
\language=\csname l@#1\endcsname
\fi
#2}}
\providecommand{\BIBdecl}{\relax}
\BIBdecl

\bibitem{interaction-composer}
D.~F. {Glas}, T.~{Kanda}, and H.~{Ishiguro}, ``Human-robot interaction design
  using interaction composer eight years of lessons learned,'' in \emph{11th
  ACM/IEEE International Conference on Human-Robot Interaction (HRI)}, 2016,
  pp. 303--310.

\bibitem{robostudio}
C.~Datta, C.~Jayawardena, I.-H. Kuo, and B.~Macdonald, ``Robostudio: A visual
  programming environment for rapid authoring and customization of complex
  services on a personal service robot,'' in \emph{IEEE/RSJ International
  Conference on Intelligent Robots and Systems.}, 2012.

\bibitem{choregraphe}
E.~{Pot}, J.~{Monceaux}, R.~{Gelin}, and B.~{Maisonnier}, ``Choregraphe: a
  graphical tool for humanoid robot programming,'' in \emph{The 18th IEEE
  International Symposium on Robot and Human Interactive Communication}, 2009,
  pp. 46--51.

\bibitem{ros}
\BIBentryALTinterwordspacing
{Stanford Artificial Intelligence Laboratory et al.}, ``Robotic operating
  system.'' [Online]. Available: \url{https://www.ros.org}
\BIBentrySTDinterwordspacing

\bibitem{interaction-blocks}
A.~Saupp\'{e} and B.~Mutlu, ``Design patterns for exploring and prototyping
  human-robot interactions,'' in \emph{Proceedings of the SIGCHI Conference on
  Human Factors in Computing Systems}, 2014.

\bibitem{slu}
R.~De~Mori, F.~Béchet, D.~Hakkani-Tur, M.~Mctear, G.~Riccardi, and G.~Tur,
  ``Spoken language understanding,'' \emph{Signal Processing Magazine, IEEE},
  vol.~25, pp. 50 -- 58, 06 2008.

\bibitem{rose2017designing}
E.~J. Rose and E.~A. Bj{\"o}rling, ``Designing for engagement: using
  participatory design to develop a social robot to measure teen stress,'' in
  \emph{Proceedings of the 35th ACM International Conference on the Design of
  Communication}, 2017, pp. 1--10.

\bibitem{ostrowski2019older}
A.~K. Ostrowski, D.~DiPaola, E.~Partridge, H.~W. Park, and C.~Breazeal, ``Older
  adults living with social robots: Promoting social connectedness in long-term
  communities,'' \emph{IEEE Robotics \& Automation Magazine}, vol.~26, no.~2,
  pp. 59--70, 2019.

\bibitem{lee2017steps}
H.~R. Lee, S.~{\v{S}}abanovi{\'c}, W.~Chang, S.~Nagata, J.~Piatt, C.~Bennett,
  and D.~Hakken, ``Steps toward participatory design of social robots: mutual
  learning with older adults with depression,'' in \emph{Proceedings of the
  ACM/IEEE international conference on human-robot interaction}, 2017.

\bibitem{ambe2019older}
A.~H. Ambe, M.~Brereton, A.~Soro, M.~Z. Chai, L.~Buys, and P.~Roe, ``Older
  people inventing their personal internet of things with the iot un-kit
  experience,'' in \emph{Proceedings of the SIGCHI Conference on Human Factors
  in Computing Systems}, 2019, pp. 1--15.

\bibitem{jelen2019older}
B.~Jelen, S.~Monsey, and K.~A. Siek, ``Older adults as makers of custom
  electronics: Iterating on craftec,'' in \emph{Extended Abstracts of the 2019
  CHI Conference on Human Factors in Computing Systems}, 2019.

\bibitem{elan}
P.~Wittenburg, H.~Brugman, A.~Russel, A.~Klassmann, and H.~Sloetjes, ``Elan: a
  professional framework for multimodality research,'' in \emph{5th
  International Conference on Language Resources and Evaluation (LREC 2006)},
  2006, pp. 1556--1559.

\bibitem{charmaz2014constructing}
K.~Charmaz, \emph{Constructing grounded theory, 2nd Ed.}\hskip 1em plus 0.5em
  minus 0.4em\relax Sages, 2014.

\bibitem{simonsen2012routledge}
J.~Simonsen and T.~Robertson, \emph{Routledge international handbook of
  participatory design}.\hskip 1em plus 0.5em minus 0.4em\relax Routledge,
  2012.

\bibitem{durick2013dispelling}
J.~Durick, T.~Robertson, M.~Brereton, F.~Vetere, and B.~Nansen, ``Dispelling
  ageing myths in technology design,'' in \emph{Proceedings of the 25th
  Australian Computer-Human Interaction Conference}, 2013.

\end{thebibliography}

\end{document}